\documentclass{ifacconf}  
\usepackage{graphicx}      
\usepackage{natbib}        
\usepackage{amsmath}
\usepackage{amssymb}
\usepackage{mathrsfs}
\usepackage{algorithm}
\usepackage{algpseudocode}
\usepackage{url}            
\begin{document}
\begin{frontmatter}

\title{Reinforcement Learning-Enhanced Control Barrier Functions for Robot Manipulators} 

\author[First]{Stephen McIlvanna} \author[First]{, Nhat Nguyen Minh} \author[First]{, Yuzhu Sun,} \author[First]{Mien Van} \author[First]{, Wasif Naeem} 
\address[First]{School of Electronics, Electrical Engineering and Computer Science, Queen’s University Belfast, Belfast, United Kingdom\\ (e-mail: m.van@qub.ac.uk)}

\begin{abstract}                
In this paper we present the implementation of a Control Barrier Function (CBF) using a quadratic program (QP) formulation that provides obstacle avoidance for a robotic manipulator arm system. CBF is a control technique that has emerged and developed over the past decade and has been extensively explored in the literature on its mathematical foundations, proof of set invariance and potential applications for a variety of safety-critical control systems. In this work we will look at the design of CBF for the robotic manipulator obstacle avoidance, discuss the selection of the CBF parameters and present a Reinforcement Learning (RL) scheme to assist with finding parameters values that provide the most efficient trajectory to successfully avoid different sized obstacles. We then create a data-set across a range of scenarios used to train a Neural-Network (NN) model that can be used within the control scheme to allow the system to efficiently adapt to different obstacle scenarios. Computer simulations (based on Matlab/Simulink) demonstrate the effectiveness of the proposed algorithm.
\end{abstract}

\begin{keyword}
Safety critical control, Control barrier function, Robotic manipulator, obstacle avoidance, Parameter selection.
\end{keyword}

\end{frontmatter}

\section{Introduction}
Safety-critical control is an important area of research that is required to bridge the gaps between `can-do' ideas from academic research and the `must-do' requirements for real-world systems. For the majority of products incorporating modern digital controllers, safety concerns are not optional. Specialised equipment or military products may be inherently unsafe but in these cases the safety aspect would be addressed by means of procedure and training. As we begin to develop the autonomous systems of the future it becomes more important to address the safety concerns directly within the controller. Taking an anthropomorphic perspective we want the autonomous robot to have the safety considerations at the very front of it's `thought-process' just as we would expect of a responsible human performing the same task.

The first (not necessarily most important) aspect of safety to consider is the safety of the system itself. Controllers can apply limits to their output that can constrain the action of effectors as designed. For optimal control strategies such as model predictive control (MPC) the limits can be defined as constraints within the optimisation effectively limiting the area the solver can seek valid solutions \cite{Khan2022}. QP Control Lyapunov Function (QP-CLF), the forerunner to QP-CBF we discuss later, can implement control limits in a similar way as in the works \cite{Ames2013, Choi2020}. Here there may be scenarios where the current system state and defined control limits conflict such that no valid control output can be found to satisfy both the stability and safety conditions within the optimisation. To address this an additional (slack) variable is typically added to the stability constraint and when a conflict arises this variable is allowed to change to increase the available solution space effectively sacrificing system stability to maintain system safety when necessary \cite{Desai2022, Ames2014}. When no conflict exists the slack variable becomes insignificant within the optimisation and the defined stability aims can be met.

The more pressing safety concern, particularly for more powerful autonomous systems, is the safety of the surrounding environment which can include the presence of other robots, equipment or humans. Obstacle avoidance has been an area of active research and has become more prominent with the emergence of self-driving cars and integrated driver assist systems common on many vehicles. For simple deterministic systems it may be possible to set predefined safety boundaries and develop rules or backup controllers that can maintain system states to a safe set but this is not realistic for implementation in most real world applications. Barrier certificates were introduced in the work \cite{Prajna2004}, barrier functions are those that produce a scalar output which rapidly increases approaching the safety limits and are used to numerically demonstrate and enforce the safety of a system \cite{Ghaffari2021}. Barrier functions were employed again within Barrier Lyapunov Function (BLF) type controllers that constrain system states. In the case of the asymmetric-BLF \cite{Tee2009}, tangent-BLF \cite{Zhang2022} or universal-BLF \cite{Jin2019} the state error is the parameter which is constrained. The integral-BLF \cite{Tee2012} can constrain the system state(s) directly, which may be more useful for certain applications. Much work has been done in this area improving convergence performance to finite \cite{Lin2022} and fixed time \cite{Jin2019}. Control Barrier Function has been developed within the past decade that builds upon the barrier certificate scheme. Here the system safety is encoded as a mathematical inequality used to constrain the solutions of an optimisation which when met keeps the system in a safe state. For optimisation type controllers, MPC or QP-CLF, this inequality can be added as an additional constraint. For any other nominal controller, even external human control \cite{Xu2018}, the CBF can be formulated as a standalone filter that will make the minimal modification to the nominal control input to keep the system safe. More recently exponential \cite{Nguyen2016} and high order CBF \cite{Tan2021, Xiao2019} have been developed to work with real world implementation on higher relative degree systems. CBF safety systems have been studied for application to vehicle \cite{Son2019, Ames2014} and robotic manipulator \cite{Singletary2022} autonomous systems. For a more complete overview on the development of CBF we direct the reader to the overview papers by researchers active in this area \cite{Ames2017, Ames2019}. The effectiveness of the CBF safety filter, however, is dependent on the selection of the parameters for CBF formulation. CBF parameter selection is a topic which does not receive much attention as it is so specific to both the dynamics of the particular application and scenarios in which it is to be used.

In paper, we propose a reinforcement learning scheme that can be used to determine the best parameters to use within the CBF formulation for different scenarios. A data-set across a range of scenarios is then created to train a Neural-Network (NN) model that can be used within the control scheme to allow the system to efficiently adapt to different obstacle scenarios. In addition, we will provide more details of the actual implementation of the CBF using Matlab/Simulink and providing our code examples.

The remainder of the paper is organized as : Section II, presents background information on the robot manipulator dynamics, CBF, NN and RL. Section III presents the safety-critical controller design and implementation. Discussion of CBF parameter selection and the proposed RL scheme is given in section IV. Then we provide simulation results in Section V. Finally section VI notes some brief conclusions on this work and our future plans in this area.

\section{Mathematical Preliminaries}
\subsection{Robotic Manipulator Dynamics and Control}
Consider the dynamics of a robot manipulator given as, 
\begin{equation}\label{eq_dynamics}
M \ddot{q} + C \dot{q}+D \dot{q}+G q=\tau
\end{equation}

Where $q \in\mathbb{R}^{n} $, $\dot{q} \in\mathbb{R}^{n}$ and $\ddot{q} \in\mathbb{R}^{n}$ are the joint position, velocity and acceleration states. $\tau \in\mathbb{R}^{n}$ is a vector of joint torque control inputs. $M \in\mathbb{R}^{n\times n}$ is the system inertia matrix for the current position states which is positive-definite. $C \in\mathbb{R}^{n}$ is a vector of Coriolis and centripetal forces dependant on the current position and velocity states. $D \in\mathbb{R}^{n}$ and $G \in\mathbb{R}^{n}$ are vectors of forces acting on the manipulator due to friction and gravity respectively.

It will be necessary to translate from joint position vector $q$ to a Cartesian position state in 3D space $\eta \in\mathbb{R}^{m}$. This can be described as a kinematic mapping transformation $\Omega : \in\mathbb{R}^{n} \mapsto \in\mathbb{R}^{m}$,
\begin{flalign}\label{eq_jacobian}
\begin{array}{ll}
\hspace{5pt} \eta = \Omega(q), \hspace{10pt} \dot{\eta} = J(q)\dot{q}, \hspace{10pt}
\ddot{\eta} = J(q)\ddot{q} + \dot{J}(q)\dot{q}
\end{array} &&
\end{flalign}
The matrix $J \in\mathbb{R}^{n \times n}$ is termed the Jacobian transformation matrix which provides the relationship between joint angle ($\dot{q}$) and end effector ($\dot{\eta}$) velocities. For this work we are concerned with the operation of the CBF filter added to prevent the end effector colliding with obstacles. A computed torque control (CTC) nominal controller is used defined as,
\begin{flalign}\label{eq_ctc}
\tau = (C\dot{q} + G + D\dot{q}) + M(\ddot{q}_d - K_d \cdot \dot{q}_{e} - K_p\cdot q_e) && 
\end{flalign}
Definitions of $M, C, D$ and $G$ remain unchanged, $\ddot{q}_d$ is the desired joint accelerations, $\dot{q}_{e} = \dot{q} - \dot{q}_d, q_e = q - q_d$ the respective joint velocity and position errors against the desired trajectory. Here we ignore friction effects thus term $D\dot{q} = 0$. $K_p$ and $K_d$ are proportional and derivative gain parameters used to tune the controller response. 
\subsection{Control Barrier Function}
A more complete overview on the foundations of CBF is given in the work \cite{Ames2019}, we present a precises of the key points used to develop the our safety function and QP in the next section. 
Given the general form of a nonlinear affine dynamical system as,
\begin{equation}\label{eq_cbfDyn}
\dot{x} = f(x) + g(x)u,
\end{equation}
with $x \in D \subset \mathbb{R}^n$ the $n$ dimension vector of states and $u \in U \subset \mathbb{R}^m$ the $m$ dimension vector of control inputs. Safety of the system is viewed as ensuring that the control input passed to the system does not cause the system states to leave a defined safe set, under the assumption that the system must begin in said safe-set. We consider a safe-set \textbf{\textit{S}} as the super-level set of a safety function $h : D \subset \mathbb{R}^n \mapsto \mathbb{R}$ we design as,
\begin{flalign}
\textbf{\textit{S}}             &= \{ x \in D \subset \mathbb{R}^n : h(x) \geq 0 \}, \nonumber\\
\partial\textbf{\textit{S}}     &= \{ x \in D \subset \mathbb{R}^n : h(x) = 0 \}, \nonumber\\
Int(\textbf{\textit{S}})        &= \{ x \in D \subset \mathbb{R}^n : h(x) > 0 \}. &&
\end{flalign}
From definition 2 of \cite{Ames2019}, $h$ is a CBF if there exists an extended class $\mathcal{K} _{\infty}$ function $\alpha$ such that for (\ref{eq_cbfDyn}),
\begin{flalign}
\sup_{u\in U}[L_f h(x) + L_g h(x)u ] \geq - \alpha(h(x)), \forall x \in D.
\end{flalign}
We deal with our second order system using the exponential CBF (ECBF) scheme described in \cite{Nguyen2016, Son2019}, omitted here for brevity.
\subsection{Neural Network Prediction Model}
\begin{figure}
\begin{center}
\includegraphics[width=7cm]{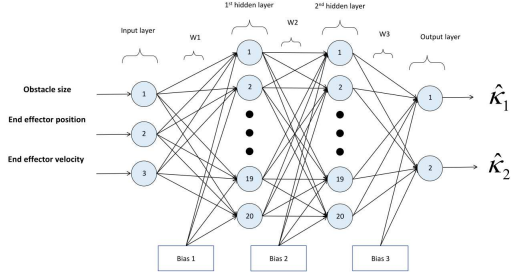}    
\caption{ANN for CBF parameter prediction.}
\label{figNN}
\end{center}
\end{figure}

Artificial Neural Networks (ANN) are now an established tool to correlate input and output data that have nonlinear relationships. Figure \ref{figNN} shows an overview of our ANN model. We initially have only considered the obstacle size input scenario but will expand this to include different relative velocities and approach conditions to deal with moving obstacles in the future. The total signals in the forward pass can be calculated as,
\begin{flalign}
y_i = W_i x + b_i, \hspace{10pt} a_i = f_i(y_i), \hspace{10pt} 
\hat{\kappa_1} = a_{31}, \hspace{10pt} \hat{\kappa_1} = a_{32}. &&
\end{flalign}
Where $W_i$ are the weight matrices and $b_i$ the bias matrices. The activation function is the \textit{tanh} function,
\begin{flalign}
f_i(y_i) = \frac{2}{(1 + e^{-2y_i})} - 1 . &&
\end{flalign}
A back-propagation algorithm is used to find the weights providing the best mapping between inputs and outputs. The cross-entropy function $J$ is a loss function given as,
\begin{flalign}
J =  \sum_{i=1}^{n} \sum_{j=1}^{2} [-\kappa_{ij} \ln{(\hat{\kappa}_{ij})} - (1 - \kappa_{ij})\ln{(1 - \hat{\kappa}_{ij})} ] &&
\end{flalign}

\subsection{Reinforcement Learning}
Reinforcement learning is a technique within the area of machine learning where the learning agent is encouraged to explore the range of actions it is able to take given an initial state. The aim is to produce the maximum reward calculated with a cost-function that assesses the resultant state from the action taken. We will initially take the approach of trying every single combination of state-action pairs to produce a complete reward map but for the majority of applications this is completely impractical due to the amount of time taken to perform every test particularly as the number of states or range within an single state increase. Then we will begin to develop a scheme that considers a rating score (reward) after each run then decides what set of parameters to try on the next iteration. This should allow the best parameter solutions to be found with less time and computational effort that will instead be directed to expanding the range of tested scenarios that can allow the CBF to adapt to the range of possible real-world scenarios.

\section{CBF Filter Design}
Consider the safety function for our end effector as the Euclidean distance between our end effector and an obstacle to be avoided. 
\begin{equation}\label{eq_hx}
h(x) = \sqrt {\left(x_{ee}-x_{o}\right)^2 + 
       \left(y_{ee}-y_{o}\right)^2 + 
       \left(z_{ee}-z_{o}\right)^2 } 
       \geq r_{m}
\end{equation}
We denote the end effector and obstacle Cartesian positions with subscript `ee' and `o' respectively. $r_m$ accounts for the distance to subtract for the obstacle and end-effector clearance radii defined as $r_m = r_{o} + r_{ee} + r_{pad}$, where $r_{o}$ defines the radius of the spherical obstacle, $r_{ee}$ a spherical clearance area around the end-effector that must avoid the obstacle and $r_{pad}$ is an additional small clearance margin we add for safety. Squaring both sides then rearranging (\ref{eq_hx}) gives,
\begin{equation}\label{eq_hx2}
h(x) = \left(x_{ee}-x_{o}\right)^2 + 
       \left(y_{ee}-y_{o}\right)^2 + 
       \left(z_{ee}-z_{o}\right)^2 -
       r_{m}^2
       \geq 0
\end{equation}
We can state that if the inequality in (\ref{eq_hx2}) holds the system state is safe at any instant. For the ECBF we now define the subsequent Lie derivatives making the substitution $sep_{\eta} = (\eta_{ee} - \eta_{o}) $ for ease of notation.
\begin{flalign}\label{eq_ld1}
L_f h(x)= 2 \left({sep}_x\right) \dot{x} + 
          2 \left({sep}_y\right) \dot{y} + 
          2 \left({sep}_z\right) \dot{z} &&
\end{flalign}

\begin{flalign}\label{eq_ld2}
\begin{array}{ll}
L_f^2 h(x)= & 2\left(sep_x\right) \ddot{x}+2(\dot{x})^2 +
              2\left(sep_y\right) \ddot{y}+2(\dot{y})^2+\\
            & 2\left(sep_z\right) \ddot{z}+2(\dot{z})^2
\end{array} &&
\end{flalign}
We substitute vector $S_n=\left[2sep_x \hspace{7pt} 2sep_y \hspace{7pt} 2sep_z \right] $, gather terms measured from state feedback in $\Gamma_n$ and substitute from system dynamics (\ref{eq_dynamics}) to give us a format suitable for use in the QP solver as follows,
\begin{flalign}\label{eq_ld22}
\begin{array}{ll}
L_f^2 h(x) &= S_n \ddot{\eta} + \Gamma_1,  
            \hspace{10pt} where \hspace{10pt} \Gamma_1 = 2\dot{x}^2 + 2\dot{y}^2 + 2\dot{z}^2 \\
            &= S_n \left( J M^{-1} \left( \tau - \Sigma_{CDG} \right) \right) + \Gamma_2, \\
            &\hspace{30pt} where \hspace{7pt} \Gamma_2 = \Gamma_1 + S_n(\dot{J}\dot{q}),\\ &\Sigma_{CDG}=C\dot{q}+D\dot{q}+Gq.
\end{array}&&
\end{flalign}
We now consider the control input $\tau$ in two parts, $\tau_{nom}$ as the fixed input from the nominal controller and $\tau_{qp}$ as the output from our QP that when subtracted from the nominal control will ensure the resultant control does not cause the system to enter an unsafe state. Rewriting (\ref{eq_ld22}),
\begin{flalign}\label{eq_ld23}
\begin{array}{ll}
L_f^2 h(x)  &= S_n \left( J M^{-1} \left( (\tau_{nom} - \tau_{qp}) - \Sigma_{CDG} \right) \right) + \Gamma_2 \\
            &= S_n \left( J M^{-1}(-\tau_{qp}) \right) + \Gamma_3 \\
            &\hspace{-10pt} where \hspace{10pt} \Gamma_3 = \Gamma_2 + S_n\left( J M^{-1} \left( \tau_{nom} - \Sigma_{CDG} \right) \right)
\end{array} &&
\end{flalign}
Separating the second Lie derivative term into the fixed nominal and computed QP parts,
\begin{flalign}\label{eq_ld24}
\begin{array}{ll}
L_f^2 h(x)  &= L_f^2 h(x)_{qp} + L_f^2 h(x)_{nom} \\
            &\hspace{-10pt} where \hspace{10pt} L_f^2 h(x)_{nom} = \Gamma_3 \\
            & \hspace{27pt}L_f^2 h(x)_{qp} \hspace{9pt}= S_n \left( J M^{-1}(-\tau_{qp}) \right)
\end{array} &&
\end{flalign}
Now constructing our ECBF and rearranging for the QP constraint format, $Ax \leq b$ where $x = -\tau_{qp}$ and any measured terms are summed in the $b$ vector,
\begin{flalign}\label{eq_cbf}
L_f^2 h(x)                   + \hat{\kappa}_2 L_f^1 h(x) + \hat{\kappa}_1 h(x) &\geq 0 \nonumber\\
L_f^2 h_{nom} + L_f^2 h_{qp} + \hat{\kappa}_2 L_f^1 h(x) + \hat{\kappa}_1 h(x) &\geq 0 &&
\end{flalign}
\begin{flalign}
-L_f^2 h_{qp}       &\leq L_f^2 h_{nom} + \hat{\kappa}_2 L_f^1 h(x) + \hat{\kappa}_1 h(x) \nonumber \\
A \cdot \tau_{qp}   &\leq b, \hspace{10pt} \textit{where } A = S_n J M^{-1} &&
\end{flalign}
Parameters $\hat{\kappa}_1, \hat{\kappa}_2$ must be estimated correctly to ensure the operation of the system remains safe. This is in a similar manner of selecting gain values for a controller scheme where choosing the wrong values can lead to poor performance and system instability, there is no equivalent to the Ziegler-Nichols method for tuning PID controllers for example. Heuristic methods are typically employed for this and it is not discussed in great detail within the literature. In the next section we want to investigate this issue.

\section{CBF Parameter Selection}
\subsection{Brute Force Approach}
\begin{figure}
\begin{center}
\includegraphics[width=8cm]{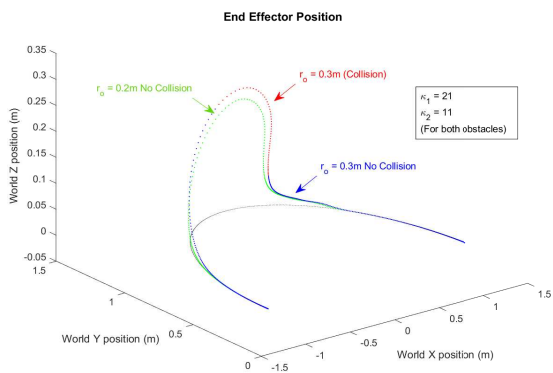}    
\caption{Resultant trajectories with different obstacles using fixed CBF parameter values.} 
\label{fig1}
\end{center}
\end{figure}
If we consider our robot manipulator arm having a fixed trajectory at a set velocity, we want to be able to avoid obstacles of varying sizes. We consider spherical obstacles with varying radius $r_o$. Figure \ref{fig1} shows the path taken by the end effector when using the same parameters for different sized objects, while it successfully avoids the smaller obstacle on the green path, the same parameters cause a collision with the larger obstacle given by the blue path. Obstacle collisions are not the only unwanted outcomes, overly conservative behaviour where the robot essentially slows or stops moving due to an upcoming obstacle or taking a very wide path around an obstacle, that may cause issues in space-constrained environments, can both occur with certain parameter selections.
Our first approach is to test in our simulated environment all the possible combinations of $r_o$, $\kappa_1$ and $\kappa_2$ parameters.
Having collected the output data of the path taken by the end effector under each of the simulated conditions we first assess a coarse success criteria which is true only if the path taken completely avoids the obstacle AND the position of the end effector is arbitrarily close to the desired trajectory by the end of the run. This second condition is necessary to counteract the aforementioned conservative behaviour where the robot does not progress along the desired trajectory in a timely manner. From the successful runs we want to apply a rating score for each run which will be a function of the total control effort and the cumulative separation from the original desired trajectory across the simulation run. We define the normalised cost function as,
\begin{flalign}\label{eq_score}
&Run(r_o, \kappa_1, \kappa_2)_{score}  = \nonumber \\
            &\hspace{30pt} goodRun \times \left(0.5\frac{Ctrl_{min}}{RunCtrl} +0.5\frac{Tsep_{min}}{RunTsep}  \right) &&
\end{flalign}
The parameter $goodRun$ is a Boolean value equal to 1 when the conditions for a good run described previously are met. $Ctrl_{min}$ and $Tsep_{min}$ are the minimum absolute total control effort and trajectory separation values from all runs, $RunCtrl$ and $Tsep$ are the values from the individual run being assessed. Using this scheme we can rank the good runs and select the highest scoring parameter set for a given obstacle size to use as data to train a neural network that can predict the best parameters based on a given obstacle size.

\subsection{Reinforcement Learning Approach}
While the previous approach covers the full range of parameters available, it is very time consuming particularly if we want to increase the range of an individual parameter or consider additional parameters as we will in future work. To address this we introduce a rudimentary RL agent to guide the search for best parameters. After each successful iteration the scores for each run are re-evaluated then the choice of the next set of parameters is made dependant on the latest best score or consecutively decreasing $Run_{score}$. We introduce an algorithm to achieve this as follows,

\begin{algorithm}
\caption{Parameter Selection For Single Scenario}\label{alg1}
\begin{algorithmic}[1]
\State initialise : $r_o = 0.2 $ \Comment{Or From Algorithm 2}
\State initialise : $\kappa_1 = \kappa_2 = 1, \hspace{5pt} s_1 = s_2 = \gamma, \hspace{5pt} flag = 0 $

\While{flag $\neq 1$}
\State RunData $\gets \boldsymbol{Run Simulation}$($r_o$, $\kappa_1$, $\kappa_2 $)
\State Update Scores \Comment{for this and all previous runs}
\If{$ \text{thisRun.score} > \text{bestScore   } $}
\State thisRun.score = bestScore
\State \textbf{goto } \textit{inc-$\kappa_2$}
\Else{}
    \If{ $i$ decreasing score $*_1$ $\|$ $\kappa_2 \geq 1.3\kappa_1$ }
        \State \textbf{goto }\textit{inc-$\kappa_1$}
    \EndIf
    \If{ $j$ decreasing score $*_2$}
    \State flag = 1, \textbf{goto } \textit{nextrun} \Comment{finished}
    \EndIf
\EndIf
\State \textit{inc-$\kappa_2 $} : $\kappa_2 = \kappa_2 + s_2$, \textbf{goto } \textit{nextrun}
\State \textit{inc-$\kappa_1 $} : $\kappa_1 = \kappa_1 + s_1, \kappa_2 = 1$
\State \textit{nextrun}
\EndWhile
\end{algorithmic}
\end{algorithm}
$*_1$ Here we assess if there have been $i$ consecutive decreasing scores within the scope of a single $\kappa_1$ parameter value.
$*_2$ Whereas here we assess if there have been $j$ consecutive decreasing scores across increasing $\kappa_1$ values. 
This process makes some assumptions on the behaviour of the parameters founded on observations when manually testing values for the CBF filter. First is that generally having $\kappa_2 > \kappa_1$ produces more conservative behaviour and in the worst case completely inhibits movement of the end effector when approaching an obstacle. Line 10 deals with this by not testing with values of $\kappa_2$ beyond a certain point. Secondly we assume we are looking for the lowest possible parameter values hence we begin at a value of $1$ for each and progressively work upwards until the resultant score of consecutive runs begins to decrease. We can test the algorithm using data generated from the `brute-force' approach to adjust constants $\gamma, i, j$ to most efficiently arrive at the best performing parameter values. The next step is to then use Algorithm \ref{alg1} iterating across different sized obstacles to produce a data-set in the format of Table \ref{table_data} that can be used to train a NN model. Within the controller this NN model can be given the obstacle size to provide estimations to the CBF filter of the most appropriate parameters to avoid the obstacle safely and efficiently.
\begin{table}
\begin{center}
\begin{tabular}{||c c l||} 
 \hline
 Inputs ($r_o)m$ & Outputs & Comments \\ [0.5ex] 
 \hline \hline
 $0.1$ & $\kappa_1(1), \kappa_2(1)$      & select top $k$ values \\ 
 $0.1$ & $\kappa_1(..), \kappa_2(..)$    & for each $r_o$ input  \\ 
 $0.1$ & $\kappa_1(k), \kappa_2(k)$      &  \\ 
 $0.2$ & $\kappa_1(1), \kappa_2(1)$      &  \\ 
 $0.2$ & $\kappa_1(2), \kappa_2(2)$      &  \\ 
 ...   & ...                             &  \\ 
 $(r_o)_{max}$ & $\kappa_1(k), \kappa_2(k)$      &  \\ 
 \hline
\end{tabular}
\caption{\label{table_data}Data generated from all simulations used to train NN parameter prediction model.}
\end{center}
\end{table}

\section{SIMULATIONS}
\subsection{Simulink CBF Model}
We provide the code used for simulations in this work via github repository \cite{sm_git_ifac23} also including video demonstrations of the parameter variations and experiments. We use the Simulink Robotics System Toolbox \cite{simulink_rst}, in particular the Jacobian Transformation, Mass-matrix, Gravity and velocity product torque blocks that generate values for the parameters discussed in the system dynamics (\ref{eq_dynamics}). The CBF QP is solved using the \textit{quadprog()} function within Matlab with console logging option suppressed to improve the simulation performance. The nominal CTC is defined by (\ref{eq_ctc}) with configurable gain parameters set as $K_p = 100, K_d = 20$. The robotic model is the Universal Robotics UR10, a 6 degree of freedom (DOF) manipulator, we target a clearance radius of $10cm$ around the end-effector setting $r_{ee} = 0.1, r_{pad} = 0.05$ in (\ref{eq_hx}). Our desired trajectory is a simple semi-circular path around the base joint at near full extension parallel to the x-y ground plane.

With the 6DOF manoeuvrability the CBF has many different options in terms of which joints it can apply correction force in order to keep the safety function inequality valid. In early tests it was noticed that the wrist joints were being affected the most to produce the necessary movement to avoid the obstacle. For situations where maintaining the pose of the end effector is critical during avoidance we can achieve a reasonably stable pose of the end-effector by inhibiting the CBF from operating on the wrist joints (three joints closest to end-effector) of the machine. This is achieved using the equality condition below added to the QP formulation,
\begin{flalign}
\text{Equality Constraint} \hspace{20pt} A_{eq}\cdot \tau = b_{eq} \nonumber \\
where \hspace{10pt} A_{eq}=\left[\begin{array}{llllll}
0 & 0 & 0 & 0 & 0 & 1 \\
0 & 0 & 0 & 0 & 1 & 0 \\
0 & 0 & 0 & 1 & 0 & 0
\end{array}\right], \hspace{5pt} 
b_{eq} = \left[0 \hspace{5pt} 0 \hspace{5pt} 0 \right]^{T} &&
\end{flalign}

\begin{figure}
\begin{center}
\includegraphics[width=8cm]{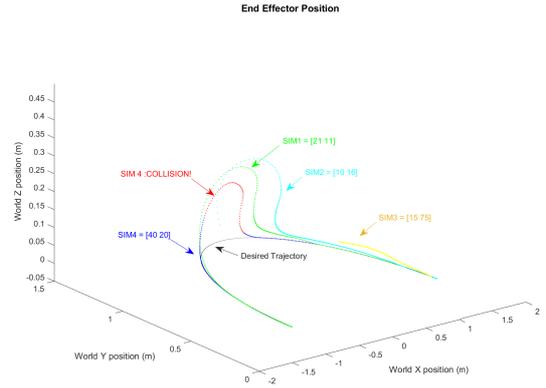}    
\caption{Resultant trajectory of four different parameter sets. $r_o = 0.2m$, parameters indicated on figure as $[\kappa_1 \hspace{5pt} \kappa_2]$}
\label{fig2}
\end{center}
\end{figure}

Figure \ref{fig2} shows the paths taken under the influence of the CBF filter with different parameter settings. Here we identify the trajectory of sim1 with the best behaviour, sim2 and sim3 show an increasing conservative action of the CBF causing the motion of the end-effector to not progress along the trajectory. In sim4 we observe a parameter selection that cause a collision with the obstacle. 
\subsection{NN prediction model}
\begin{figure}
\begin{center}
\includegraphics[width=8cm]{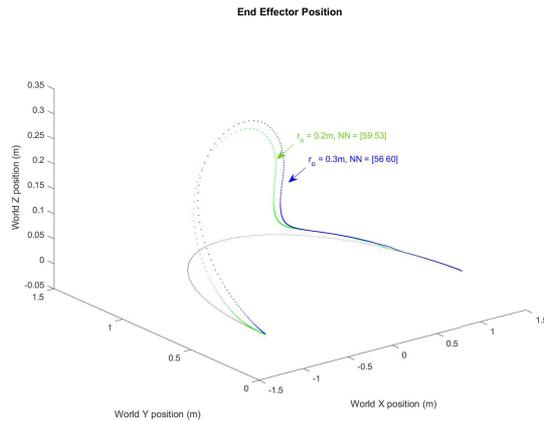}    
\caption{Resultant trajectory for two different obstacles, parameters provided from NN indicated on figure as $[\kappa_1 \hspace{5pt} \kappa_2]$} 
\label{fig3}
\end{center}
\end{figure}

We generated a data-set using a limited version of the `brute-force' scheme, with parameter sets $r_o = [0.05,0.1,0.2,0.25,0.3,0.4,0.5,0.6] (m)$,\\ $\kappa_1, \kappa_2 = [1, 3, 5, 10, 15, 25, 30, 40, 50, 60, 70, 80, 100]$ for all combinations. Performing each simulation takes an average of $20seconds$ for 1764 runs in a total time of around 10 hours on this reduced set using Matlab 2021b with an i7-1165G7 CPU. After conducting these simulations we applied the scoring metrics (\ref{eq_score}), then selected the top 5 scoring runs from each obstacle size. Training a NN model in Matlab then adding this model as a block within Simulink, passing to it the size of the obstacle, it was able to provide the CBF with parameters to successfully avoid two different sized obstacles as shown in Figure \ref{fig3}. For the RL scheme we have done some initial validation on a larger data-set generated with one obstacle size with a larger range of $\kappa_1, \kappa_2$ parameter values. This has shown some initial success in being able to correctly get to the best scoring parameters for a single scenario in a much faster run time. The early results with this work shows some promise and we intend to continue to implement the RL scheme in order to be able to generate larger data-sets and improve the prediction capabilities of our model which we hope to present in the future.

\section{Conclusion}
In this paper we have developed a CBF filter and using a RL method to test the various parameter options ultimately being able to create a prediction model that can provide parameters to a robotic manipulator system during operation when it encounters an obstacle it must avoid. We hope to make further progress on this scheme in the near future, this will include generating larger data-sets using the RL algorithm, testing on a wider range of obstacle sizes, considering moving obstacles and the speed of the given desired trajectory as another input variable to our prediction model. Gaussian process is another approach for creating a prediction model that we will explore due to its ability to produce good approximations with smaller data-sets and the confidence interval metrics it provides.

\bibliography{ifacconf}                            

\begin{thebibliography}{22}
\providecommand{\natexlab}[1]{#1}
\providecommand{\url}[1]{\texttt{#1}}
\providecommand{\urlprefix}{URL }
\expandafter\ifx\csname urlstyle\endcsname\relax
  \providecommand{\doi}[1]{doi:\discretionary{}{}{}#1}\else
  \providecommand{\doi}{doi:\discretionary{}{}{}\begingroup
  \urlstyle{rm}\Url}\fi

\bibitem[{Ames et~al.(2019)Ames, Coogan, Egerstedt, Notomista, Sreenath, and
  Tabuada}]{Ames2019}
Ames, A.D., Coogan, S., Egerstedt, M., Notomista, G., Sreenath, K., and
  Tabuada, P. (2019).
\newblock Control barrier functions: Theory and applications.
\newblock \emph{2019 18th European Control Conference (ECC)}, 3420--3431.
\newblock \doi{10.23919/ECC.2019.8796030}.

\bibitem[{Ames et~al.(2014)Ames, Grizzle, and Tabuada}]{Ames2014}
Ames, A.D., Grizzle, J.W., and Tabuada, P. (2014).
\newblock Control barrier function based quadratic programs with application to
  adaptive cruise control.
\newblock \emph{53rd IEEE Conference on Decision and Control}, 6271--6278.
\newblock \doi{10.1109/CDC.2014.7040372}.
\newblock \urlprefix\url{http://ieeexplore.ieee.org/document/7040372/}.

\bibitem[{Ames and Powell(2013)}]{Ames2013}
Ames, A.D. and Powell, M. (2013).
\newblock Towards the unification of locomotion and manipulation through
  control lyapunov functions and quadratic programs.
\newblock 219--240.
\newblock \doi{10.1007/978-3-319-01159-2_12}.
\newblock
  \urlprefix\url{http://link.springer.com/10.1007/978-3-319-01159-2_12}.

\bibitem[{Ames et~al.(2017)Ames, Xu, Grizzle, and Tabuada}]{Ames2017}
Ames, A.D., Xu, X., Grizzle, J.W., and Tabuada, P. (2017).
\newblock Control barrier function based quadratic programs for safety critical
  systems.
\newblock \emph{IEEE Transactions on Automatic Control}, 62, 3861--3876.
\newblock \doi{10.1109/TAC.2016.2638961}.

\bibitem[{Choi et~al.(2020)Choi, Castañeda, Tomlin, and Sreenath}]{Choi2020}
Choi, J., Castañeda, F., Tomlin, C.J., and Sreenath, K. (2020).
\newblock Reinforcement learning for safety-critical control under model
  uncertainty, using control lyapunov functions and control barrier functions.
\newblock \urlprefix\url{http://arxiv.org/abs/2004.07584}.

\bibitem[{Desai and Ghaffari(2022)}]{Desai2022}
Desai, M. and Ghaffari, A. (2022).
\newblock Clf-cbf based quadratic programs for safe motion control of
  nonholonomic mobile robots in presence of moving obstacles.
\newblock 16--21. IEEE.
\newblock \doi{10.1109/AIM52237.2022.9863392}.

\bibitem[{Ghaffari(2021)}]{Ghaffari2021}
Ghaffari, A. (2021).
\newblock Modular safety control for mobile robots using barrier certificates
  and modified feedback.
\newblock 4243--4248. IEEE.
\newblock \doi{10.23919/ACC50511.2021.9482636}.
\newblock \urlprefix\url{https://ieeexplore.ieee.org/document/9482636/}.

\bibitem[{Jin(2019)}]{Jin2019}
Jin, X. (2019).
\newblock Adaptive fixed-time control for mimo nonlinear systems with
  asymmetric output constraints using universal barrier functions.
\newblock \emph{IEEE Transactions on Automatic Control}, 64, 3046--3053.
\newblock \doi{10.1109/TAC.2018.2874877}.

\bibitem[{Khan et~al.(2022)Khan, Guivant, and Li}]{Khan2022}
Khan, S., Guivant, J., and Li, X. (2022).
\newblock Design and experimental validation of a robust model predictive
  control for the optimal trajectory tracking of a small-scale autonomous
  bulldozer.
\newblock \emph{Robotics and Autonomous Systems}, 147.
\newblock \doi{10.1016/j.robot.2021.103903}.

\bibitem[{Lin et~al.(2022)Lin, Chen, and Li}]{Lin2022}
Lin, X., Chen, C.C., and Li, S. (2022).
\newblock Finite-time output feedback stabilization for a class of
  output-constrained planar switched systems.
\newblock \emph{IEEE Transactions on Circuits and Systems II: Express Briefs},
  69, 164--168.
\newblock \doi{10.1109/TCSII.2021.3077603}.

\bibitem[{Mathworks(2022)}]{simulink_rst}
Mathworks (2022).
\newblock Simulink rst.
\newblock \urlprefix\url{https://uk.mathworks.com/products/robotics.html}.

\bibitem[{Nguyen and Sreenath(2016)}]{Nguyen2016}
Nguyen, Q. and Sreenath, K. (2016).
\newblock Exponential control barrier functions for enforcing high
  relative-degree safety-critical constraints.
\newblock volume 2016-July, 322--328. Institute of Electrical and Electronics
  Engineers Inc.
\newblock \doi{10.1109/ACC.2016.7524935}.

\bibitem[{Prajna and Jadbabaie(2004)}]{Prajna2004}
Prajna, S. and Jadbabaie, A. (2004).
\newblock Safety verification of hybrid systems using barrier certificates.
\newblock 477--492.
\newblock \doi{10.1007/978-3-540-24743-2_32}.
\newblock
  \urlprefix\url{http://link.springer.com/10.1007/978-3-540-24743-2_32}.

\bibitem[{Singletary et~al.(2022)Singletary, Kolathaya, and
  Ames}]{Singletary2022}
Singletary, A., Kolathaya, S., and Ames, A.D. (2022).
\newblock Safety-critical kinematic control of robotic systems.
\newblock \emph{IEEE Control Systems Letters}, 6, 139--144.
\newblock \doi{10.1109/LCSYS.2021.3050609}.

\bibitem[{S.McIlvanna(2022)}]{sm_git_ifac23}
S.McIlvanna (2022).
\newblock Simulink files for ur10 cbf simulation.
\newblock \urlprefix\url{https://github.com/smcilvanna/cbf_ur10_ifac23}.

\bibitem[{Son and Nguyen(2019)}]{Son2019}
Son, T.D. and Nguyen, Q. (2019).
\newblock Safety-critical control for non-affine nonlinear systems with
  application on autonomous vehicle.
\newblock volume 2019-December, 7623--7628. Institute of Electrical and
  Electronics Engineers Inc.
\newblock \doi{10.1109/CDC40024.2019.9029446}.

\bibitem[{Tan et~al.(2021)Tan, Cortez, and Dimarogonas}]{Tan2021}
Tan, X., Cortez, W.S., and Dimarogonas, D.V. (2021).
\newblock High-order barrier functions: Robustness, safety and
  performance-critical control.
\newblock \emph{IEEE Transactions on Automatic Control}, 1--1.
\newblock \doi{10.1109/TAC.2021.3089639}.

\bibitem[{Tee and Ge(2012)}]{Tee2012}
Tee, K.P. and Ge, S.S. (2012).
\newblock Control of state-constrained nonlinear systems using integral barrier
  lyapunov functionals.
\newblock 3239--3244.
\newblock \doi{10.1109/CDC.2012.6426196}.

\bibitem[{Tee et~al.(2009)Tee, Ge, and Tay}]{Tee2009}
Tee, K.P., Ge, S.S., and Tay, E.H. (2009).
\newblock Barrier lyapunov functions for the control of output-constrained
  nonlinear systems.
\newblock \emph{Automatica}, 45, 918--927.
\newblock \doi{10.1016/j.automatica.2008.11.017}.

\bibitem[{Xiao and Belta(2019)}]{Xiao2019}
Xiao, W. and Belta, C. (2019).
\newblock Control barrier functions for systems with high relative degree.
\newblock volume 2019-December, 474--479. Institute of Electrical and
  Electronics Engineers Inc.
\newblock \doi{10.1109/CDC40024.2019.9029455}.

\bibitem[{Xu and Sreenath(2018)}]{Xu2018}
Xu, B. and Sreenath, K. (2018).
\newblock Safe teleoperation of dynamic uavs through control barrier functions.
\newblock 7848--7855. Institute of Electrical and Electronics Engineers Inc.
\newblock \doi{10.1109/ICRA.2018.8463194}.

\bibitem[{Zhang et~al.(2022)Zhang, Liu, Tang, Hou, and Wang}]{Zhang2022}
Zhang, L., Liu, H., Tang, D., Hou, Y., and Wang, Y. (2022).
\newblock Adaptive fixed-time fault-tolerant tracking control and its
  application for robot manipulators.
\newblock \emph{IEEE Transactions on Industrial Electronics}, 69, 2956--2966.
\newblock \doi{10.1109/TIE.2021.3070494}.
\newblock \urlprefix\url{https://ieeexplore.ieee.org/document/9398567/}.

\end{thebibliography}
\end{document}